\documentclass{article}

\usepackage{amsmath}
\usepackage{amssymb}
\usepackage{graphicx}
\usepackage{theorem}
\usepackage{tabularx}
\usepackage{ifthen}
\usepackage{subfigure}
\usepackage{url}

\usepackage[french]{babel}
\usepackage[utf8]{inputenc}

\newtheorem{definition}{Définition}

\newcommand{\ie}{\textit{i.e.}, }
\newcommand{\eg}{\textit{e.g.}, }

\newcommand{\name}{\textit{name}}
\newcommand{\precond}{\textit{precond}}
\newcommand{\effects}{\textit{effects}}
\newcommand{\operators}{\textit{operators}}
\newcommand{\belief}{\textit{belief}}

\newcommand{\cons}[1]{\textsf{\small{#1}}}
\newcommand{\var}[1]{\textsf{\small{?#1}}}
\newcommand{\pre}[2]{%
  \ifthenelse{\equal{#2}{}}{%
    \textsf{\small{#1}}}{%
    \textsf{\small{#1}}{\small{(#2)}}}}
\newcommand{\op}[2]{%
  \ifthenelse{\equal{#2}{}}{%
    \textsf{\small{#1}}}{%
    \textsf{\small{#1}}{\small{(#2)}}}}
\newcommand{\operator}[4][x]{%
  \ifthenelse{\equal{#1}{x}}{
    \begin{flushleft}
      \begin{tabularx}{\textwidth}[!h]{lp{100mm}}
        \multicolumn{2}{l}{\small{#2}} \\
        \hspace{5mm} precond: & \small{#3} \\
        \hspace{5mm} effects: & \small{#4} \\
      \end{tabularx}
    \end{flushleft}} {
    \begin{flushleft}
      \begin{tabularx}{\textwidth}[!h]{lp{100mm}}
        \multicolumn{2}{p{130mm}}{\textit{\textsf{\small{;; #1}}}} \\
        \multicolumn{2}{l}{\small{#2}} \\
        \hspace{5mm} precond: & \small{#3} \\
        \hspace{5mm} effects: & \small{#4} \\
      \end{tabularx}
    \end{flushleft}}}

\theoremstyle{plain}
\theorembodyfont{\rmfamily}
\newtheorem{example}{Exemple}[section]

\title{Planification distribuée par fusions \\ incrémentales de graphes}

\author{Damien Pellier, Ilias Belaidi\\
Université Paris Descartes \\
Centre de Recherche en Informatique \\
45, rue des Saints Père \\
F-75270 Paris cedex 06 \\
\texttt{Damien.Pellier@univ-paris5.fr} \\ \texttt{Ilias.Belaidi@univ-paris5.fr}
}

\date{Juin 2008}

\begin{document}

\maketitle

\begin{abstract}
Dans cet article, nous proposons un modèle générique et original pour la synthèse distribuée de plans par un groupe d'agents, appelé {\it planification distribuée par fusions incrémentales de graphes}. Ce modèle unifie de manière élégante les différentes phases de la planification distribuée au sein d'un même processus. Le modèle s'appuie sur les graphes de planification, utilisé en planification mono-agent, pour permettre aux agents de raisonner et sur une technique de satisfaction de contraintes pour l'extraction et la coordination des plans individuels. L'idée forte du modèle consiste à intégrer au plus tôt, \ie au sein du processus local de planification, la phase de coordination. L'unification de ces phases permet ainsi aux agents de limiter les interactions négatives entre leurs plans individuels, mais aussi, de prendre en compte leurs interactions positives, \ie d'aide ou d'assistance, lors de l'extraction de leurs plans individuels.
\end{abstract}

\section{Introduction}

La planification multi-agent est une problématique centrale de l'intelligence artificielle distribuée. Les applications qui peuvent bénéficier directement des avancées de ce domaine de recherche sont nombreuses, \eg la robotique \cite{alami:98}, les services web \cite{wu:03}, dans lesquelles les approches centralisées sont difficilement envisageables. Classiquement, la planification multi-agent se définit comme un processus  distribué qui prend en entrée: une modélisation des actions réalisables par des agents; une description de l'état du monde connu des agents et un objectif. Le processus doit alors retourner un ensemble organisé d'actions, \ie un plan, qui s'il est exécuté, permet d'atteindre l'objectif. Une telle définition permet facilement d'appréhender ce qu'est la planification, mais cache la complexité et le découpage des différentes phases du processus. La littérature distingue cinq phases:
\begin{enumerate}
\item {\em Une phase de décomposition.} Les agents raffinent l'objectif initial en sous objectifs atteignables individuellement par les agents. Les approches proposées reposent le plus souvent sur des techniques centralisées de planification, \eg {\sc Htn} \cite{younes:03} ou sur les graphes de planification \cite{iwen:02}.
\item {\em Une phase de délégation.} Les agents s'assignent mutuellement les objectifs obtenus lors de la première phase. Cette étape s'appuie sur des techniques d'allocation de ressources dans un contexte centralisé \cite{bar-noy:01} ou dans un contexte distribué basées par exemple sur la négociation \cite{zlotkin:90}, l'argumentation \cite{tambe:99}, ou encore la synchronisation \cite{clement:03}.
\item {\em Une phase de planification locale.} Au cours de cette phase, chaque agent essaie de trouver un plan individuel pour atteindre l'objectif qui lui est assigné à l'étape 2. Les techniques utilisées reposent sur les algorithmes classiques de planification mono-agent, \eg \cite{blum:97,nau:03,penberthy:92}.
\item {\em Une phase de coordination.} Lors de cette phase chaque agent vérifie que le plan local élaboré à  l'étape précédente n'entre pas en conflit avec les plans locaux des autres agents. Contrairement à la planification mono-agent, cette étape est nécessaire pour garantir l'intégrité fonctionnelle du système, \ie assurer que le but de chaque agent reste atteignable dans le contexte global. La problématique de la coordination a été tout particuliérement étudiée dans le domaine des systèmes multi-agents notamment en la considérant comme un processus de fusion de plans \cite{tonino:02} ou de résolution de conflits \cite{cox:05}.
\item {\em Une phase d'exécution.} Finalement, le plan exempt de conflit est exécuté par les agents.
\end{enumerate}

Bien que cette décomposition soit conceptuellement correcte, l'expérience montre que les cinq phases sont souvent entrelacées: ({\it i}) les phases 1 et 2 sont souvent fusionnées à cause du lien fort qu'il existe entre les capacités des agents et la décomposition de l'objectif; ({\it ii}) la dernière phase d'exécution conduit presque toujours à une replanification \cite{fox:06} due à un événement non prévu lors de l'élaboration du plan et nécessite un retour aux phases 1, 2 et 3; ({\it iii}) la phase de coordination peut être réalisée avant \cite{shoham:95}, après \cite{tonino:02} ou pendant \cite{lesser:04} l'étape de planification locale. Cette remarque fait clairement apparaître un manque de modèles capables d'embrasser les différentes phases du processus de planification multi-agent.

Pour tenter de répondre en partie à cette critique, nous présentons dans cet article un modèle de planification multi-agent dans lequel l'élaboration d'un plan partagé est vu comme un processus collaboratif et itératif, intégrant les quatre premières étapes décrites précédemment. L'idée forte du modèle proposé consiste à amalgamer au plus tôt, \ie au sein du processus local de planification, la phase de coordination en s'appuyant sur une structure classique dans le domaine de planification, les graphes de planification \cite{blum:97}. Cette représentation possède trois principaux avantages: elle est calculable en temps polynomiale; elle permet d'utiliser les nombreux travaux portant sur la recherche d'un plan solution dans les graphes de planification; finalement, elle est assez générique pour représenter une certaine forme de concurrence entre les actions. Ce dernier point est un atout majeur dans le cadre de la planification multi-agent.

L'intégration des phases de coordination et de planification locale, au sein d'un même processus que nous appelons {\em fusions incrémentales de graphes}, présente l'avantage de limiter les interactions négatives entre les plans individuels des agents, mais aussi et surtout, de prendre en compte les interactions positives, \ie d'aide ou d'assistance entre agents. Supposons un instant que le but d'un agent soit de prendre un objet dans une pièce dont la porte est fermée à clé. Même si l'agent possède un plan pour atteindre son but une fois la porte ouverte mais qu'il n'est pas capable d'ouvrir la porte, le processus de planification échoue. Supposons maintenant qu'un second agent soit capable d'ouvrir la porte, il existe alors un plan solution. Faut-il encore que le premier agent soit en mesure de considérer, au moment de la synthèse de son plan local, l'aide que peut lui apporter le second agent, et que ce dernier soit en mesure de proposer l'action requise.

Dans la suite de l'article, nous présentons tout d'abord les définitions préliminaires nécessaires à la formalisation de notre approche, puis nous détaillons les cinq phases de l'algorithme de planification distribuée par fusions incrémentales de graphes: la décomposition du but global en buts individuels; l'expansion et la fusion des graphes de planification; l'extraction d'un plan individuel et finalement la coordination des plans individuels solution.

\section{Définitions préliminaires}

La syntaxe et la sémantique utilisées dans notre approche s'appuient sur la logique du premier ordre, à savoir, les constantes, les prédicats et les variables. Le nombre des symboles de prédicats et de constantes est fini. Les propositions sont des tuples d'arguments, \ie des constantes ou des variables, négatives ou positives, \eg \pre{at}{\var{x}, \cons{paris}}, ou $\neg$\pre{at}{\var{x},\cons{paris}}. Pour les prédicats et les constantes, nous utiliserons des chaînes de caractères alphanumériques comportant au moins deux caractères. Les variables seront définies par un seul caractère précédé par {\small \textsf{?}}. Nous définissons la co-désignation comme la relation d'équivalence entre les constantes et les variables. Par extension, deux propositions se co-désignent si elles sont toutes deux négatives ou positives, si elles possèdent le même nombre d'arguments et si tous leurs arguments se co-désignent deux à deux, \eg \pre{at}{\var{x}, \cons{paris}} et \pre{at}{\var{y}, \cons{paris}} se co-désignent si $\var{x} = \var{y}$.

Les opérateurs de planification sont définis comme des fonctions de transition au sens {\sc Strips}. Nous nous plaçons ici dans le cadre de la planification classique: actions instantanées, statiques, déterministes et observabilité totale.

\begin{definition}[Opérateur]
Un {\em opérateur} $o$ est un triplet de la forme $(\name(o),$ $\precond(o),$ $\effects(o))$ ou $\name(o)$ définit le nom de l'opérateur, $\precond(o)$ l'ensemble des préconditions de $o$ à satisfaire et $\effects(o)$ l'ensemble de ses effets. Par la suite nous noterons respectivement $\effects^{+}(o)$ les effets positifs et $\effects^{-}(o)$ les effets négatifs d'un opérateur $o$.
\end{definition}

\begin{definition}[Action]
Une {\em action} est une instance d'un opérateur. Si $a$ est une action et $s_i$ un état tel que $\precond(a) \subseteq s_i$ alors l'action $a$ est {\em applicable} dans $s_i$ et le résultat de son application est un état:  $s_{i+1} = (s_{i} - \effects^{-}(a)) \cup \effects^{+}(a)$.
\end{definition}

\begin{definition}[Indépendance]
Deux actions $a$ et $b$ sont {\em indépendantes}\footnote{L'indépendance entre actions  n'est pas une propriété spécifique à un problème de planification donné. Elle découle uniquement de la description des opérateurs des agents.} ssi $\effects^{-}(a)$ $\cap (\precond(b) \cup \effects^{+}(b)) = \emptyset$ et $\effects^{-}(b) \cap (\precond(a) \cup \effects^{+}(a)) = \emptyset$. Par extension un ensemble $A$ d'actions est indépendant si toutes les paires d'actions de $A$ sont indépendantes deux à deux.
\end{definition}

\begin{definition}[Graphe de planification]
Un {\em graphe de planification} ${\cal{G}}$ est un graphe orienté par niveaux\footnote{Un graphe dont les sommets sont partitionnés en ensembles disjoints tels que les arcs connectent seulement les sommets des niveaux adjacents.} organisés en une séquence alternée de propositions et d'actions de la forme $\langle P_0, A_0, P_1, \ldots, A_{i-1}, P_i \rangle$. $P_0$ contient les propositions de l'état initial. Les niveaux $A_i$ avec $i > 0$ sont composés de l'ensemble des actions applicables à partir des niveaux propositionnels $P_i$ et les niveaux $P_{i+1}$ sont constitués des propositions produites par les actions de $A_i$. Les arcs sont définis de la manière suivante: pour toutes les actions $a \in A_i$, un arc relie $a$ avec toutes les propositions $p \in P_i$ qui représentent les préconditions de $a$; pour toutes les propositions $p \in P_{i+1}$ un arc relie $p$ aux actions $a \in A_i$ qui produisent ou suppriment $p$. À chaque niveau le maintien des propositions du niveau $i$ au niveau $i + 1$ est assuré par des actions appelées \op{no-op}{} qui ont une unique proposition comme précondition et effet positif.
\label{Def:Planning-Graph}
\end{definition}

\begin{definition}[Exclusion mutuelle]
Deux actions $a$ et $b$ sont {\em mutuellement exclusives} ssi $a$ et $b$ sont dépendantes ou si une précondition de $a$ est mutuellement exclusive avec une précondition de $b$. Deux propositions $p$ et $q$ d'un niveau propositionnel $P_i$ d'un graphe de planification sont mutuellement exclusives si toutes les actions de $A_i$ avec $p$ comme un effet positif sont mutuellement exclusives avec toutes les actions qui produisent $q$ au même niveau, et s'il n'existe aucune action de $A_i$ qui produit à la fois $p$ et $q$. Par la suite, nous noterons l'ensemble des exclusions mutuelles d'un niveau $A_i$ et $P_i$ respectivement $\mu A_i$ et $\mu P_i$.
\end{definition}

\begin{definition}[Point fixe]
Un graphe de planification $\cal{G}$ de niveau $i$ a atteint son {\em point fixe} si $\forall i, i > j,$ le niveau $i$ de $\cal{G}$ est identique au niveau $j$, \ie $P_{i} = P_j$, $\mu P_{i} = \mu P_j$, $A_{i} = A_j$ et $\mu A_{i} = \mu A_i$.
\end{definition}

\begin{definition}[Agent]
Un {\em agent} est un tuple $\alpha =$ $(\name(\alpha),$ $\operators(\alpha),$ $\belief(\alpha))$ où $\name(\alpha)$ définit le nom de l'agent, $\operators(\alpha)$ l'ensemble des opérateurs que peut réaliser l'agent, et  $\belief(\alpha)$ les croyances de l'agent.
\end{definition}

De plus, nous introduisons $\precond(\alpha)$ qui définit l'ensemble des propositions utilisées dans la description des préconditions des opérateurs de $\alpha$,  $\effects^{+}(\alpha)$ et $\effects^{-}(\alpha)$ qui représentent respectivement l'ensemble des propositions utilisées dans la description des effets négatifs et positifs des opérateurs de $\alpha$. Nous supposons que ces trois ensembles constituent la partie publique d'un agent, \ie les agents peuvent accéder à ces informations  à n'importe quel moment du processus de planification. En d'autres termes, en rendant publique ces informations, un agent publie les propriétés du monde qu'il est en mesure de modifier.

\begin{definition}[Interaction négative]
Une action $a$ d'un graphe de planification d'un agent $\alpha$ {\em menace} l'activité d'un agent $\beta$ ssi une proposition $p \in \effects^{-}(a)$ co-désigne une proposition $q$ telle que $q \in \precond(\beta)$ ou $q \in \effects^{+}(\beta)$.
\label{Def:Relation-Negative}
\end{definition}

\begin{definition}[Interaction positive]
Une action $a$ d'un graphe de planification d'un agent $\alpha$ {\em assiste} l'activité d'un agent $\beta$ ssi une proposition $p \in \effects^{+}(a)$ co-désigne une proposition $q$ telle que $q \in \precond(\beta)$ ou $q \in \effects^{+}(\beta)$.
\label{Def:Relation-Positive}
\end{definition}

Notons que ces deux définitions revêtent un caractère particulièrement important dans le cadre de notre modèle. C'est en effet, sur la base de ces interactions, que les agents seront capables de planifier en intégrant les activités des autres agents.

\begin{definition}[Problème]
Un {\em problème de planification multi-agent} est un tuple ${\cal{P}} = ({\cal{A}}, g)$ où $\cal{A}$ représente l'ensemble des agents devant résoudre le problème et $g$ le but à atteindre, \ie l'ensemble des propositions qui doivent être satisfaites par les agents après l'exécution du plan. Nous faisons l'hypothèse restrictive que l'union des croyances des agents d'un problème de planification est cohérente (cas classique de la planification mono-agent), \ie pour deux agents $\alpha$ et $\beta$ $\in {\cal{A}}$, si une proposition $p \in \belief(\alpha)$ alors $\neg p \notin \belief(\beta)$. Cependant, aucune hypothèse n'est faite sur le possible partage de croyances entre les agents en termes de faits ou d'opérateurs.
\end{definition}

\begin{definition}[But individuel]
Le but individuel $g_\alpha$ d'un agent $\alpha$ pour un problème de planification multi-agent ${\cal{P}} = ({\cal{A}}, g)$ avec $\alpha \in {\cal{A}}$ est défini par $g_\alpha = \{ p \in g \ | \ p$ co-designe une proposition $q \in \effects^{+}(\alpha)\}$. Autrement dit, un agent ne peut accepter comme but que les propositions qui sont définis comme des effets des actions qu'il peut exécuter.
\label{Def:But-Individuel}
\end{definition}

\begin{definition}[Plan]
Un {\em plan} $\pi = \langle A_0,...,A_n\rangle$ est une séquence finie d'ensembles d'actions avec $n \geq 0$. Si $n = 0$, alors le plan $\pi$ est le plan vide, noté $\langle\rangle$. Si les ensembles d'actions composant le plan sont des singletons, \ie $A_0 = \{a_0\},\ldots,A_n = \{a_n\}$, nous utiliserons la notation simplifiée $\pi = \langle a_0,...,a_n\rangle$.
\end{definition}

\begin{definition}[Plan solution individuel]
Un plan $\pi_\alpha = \langle A_{0}^{\alpha},...,A_{n}^{\alpha}\rangle$ est un {\em plan solution individuel} d'un problème de planification multi-agent ${\cal{P}} = ({\cal{A}}, g)$ pour un agent $\alpha \in {\cal{A}}$ ssi tous les ensembles d'actions $A_{i}^{\alpha}$ sont indépendants et l'application des $A_{i}^{\alpha}$ à partir de $\belief(\alpha)$ définit une séquence d'états $\langle s_0, \ldots, s_n\rangle$ telle que $g_\alpha \subseteq s_n$.
\end{definition}

\begin{definition}[Plan solution global]
Un plan $\Pi = \langle A_0,...,A_n\rangle$ est un {\em plan solution global} d'un problème de planification multi-agent ${\cal{P}} = ({\cal{A}}, g)$ ssi il existe un sous-ensemble d'agents  ${\cal{A}}'$ avec ${\cal{A}}' \subseteq {\cal{A}}$  tel que l'union des buts individuels $g_\alpha$ des agents $\alpha \in {\cal{A}}'$ est égale à $g$, que tous les agents $\alpha \in {\cal{A}}'$ possèdent un plan solution individuel $\pi_\alpha = \langle A_{0}^{\alpha},...,A_{n}^{\alpha}\rangle$ et que tous les ensembles d'actions $A_i = \bigcup A_{i}^{\alpha}$  sont indépendants.
\end{definition}

\begin{example}[Dockers]
Considérons un exemple simple avec trois agents: deux agents capables de charger et de décharger des conteneurs et, un troisième qui assure le déplacement d'un conteneur d'un endroit à un autre. Le premier docker \cons{ag1} est à l'emplacement \cons{l1}. Le second docker est à l'emplacement \cons{l2}. Le but du problème est $g = \{\pre{at}{\cons{c1},\cons{l2}}, \pre{at}{\cons{c2},\cons{l1}}, \pre{at}{\cons{t1},\cons{l2}}, \pre{at}{\cons{t2},\cons{l1}}\}$. Nous donnons ci-dessous la définition des opérateurs du problème ainsi que son état initial (cf. {\it fig.} \ref{Fig:Example}):

\begin{figure}[!]
\begin{center}
\includegraphics[scale=0.3]{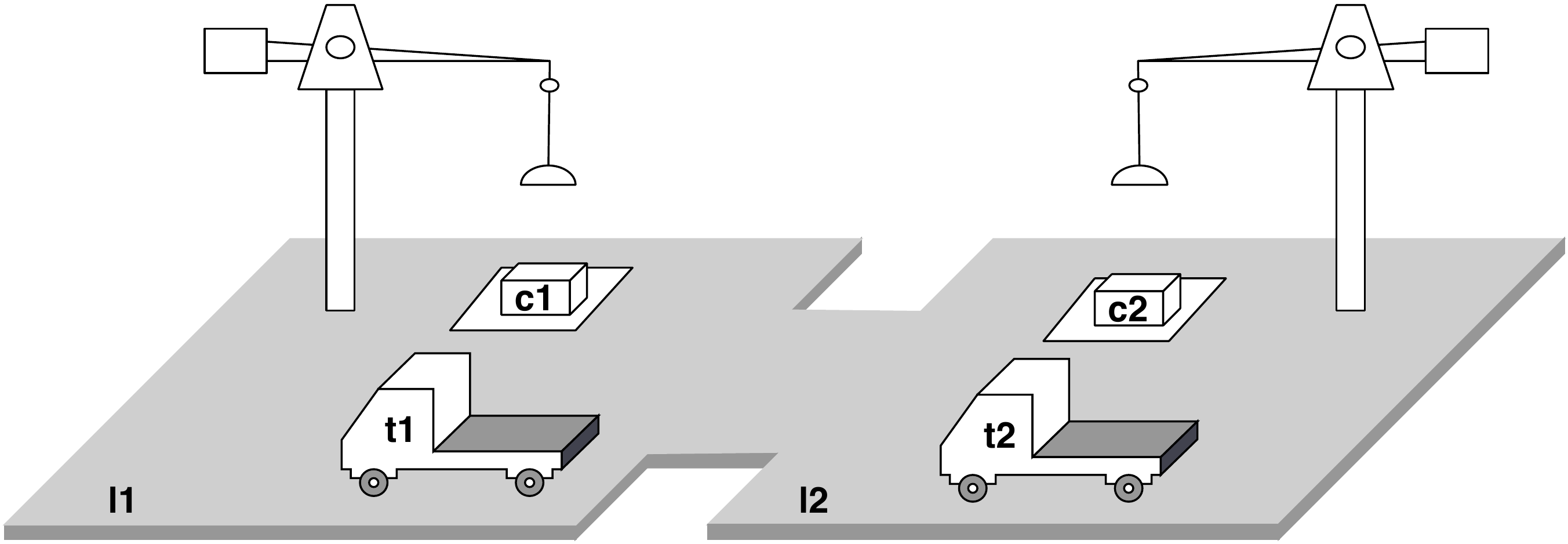}
\end{center}
\caption{État initial de l'exemple:}
\begin{center}
$\belief(\cons{ag1}) = \{\pre{at}{\cons{c1},\cons{l1}}, \pre{at}{\cons{t1}, \cons{l1}}\}$\\
$\belief(\cons{ag2}) = \{\pre{at}{\cons{c2},\cons{l2}}, \pre{at}{\cons{t2}, \cons{l2}}\}$ \\
$\belief(\cons{ag3}) = \{\pre{at}{\cons{t1},\cons{l1}}, \pre{at}{\cons{t2},\cons{l2}}\}$
\end{center}
\label{Fig:Example}
\end{figure}

\begin{figure*}[!]
\begin{center}
\includegraphics[scale=0.46]{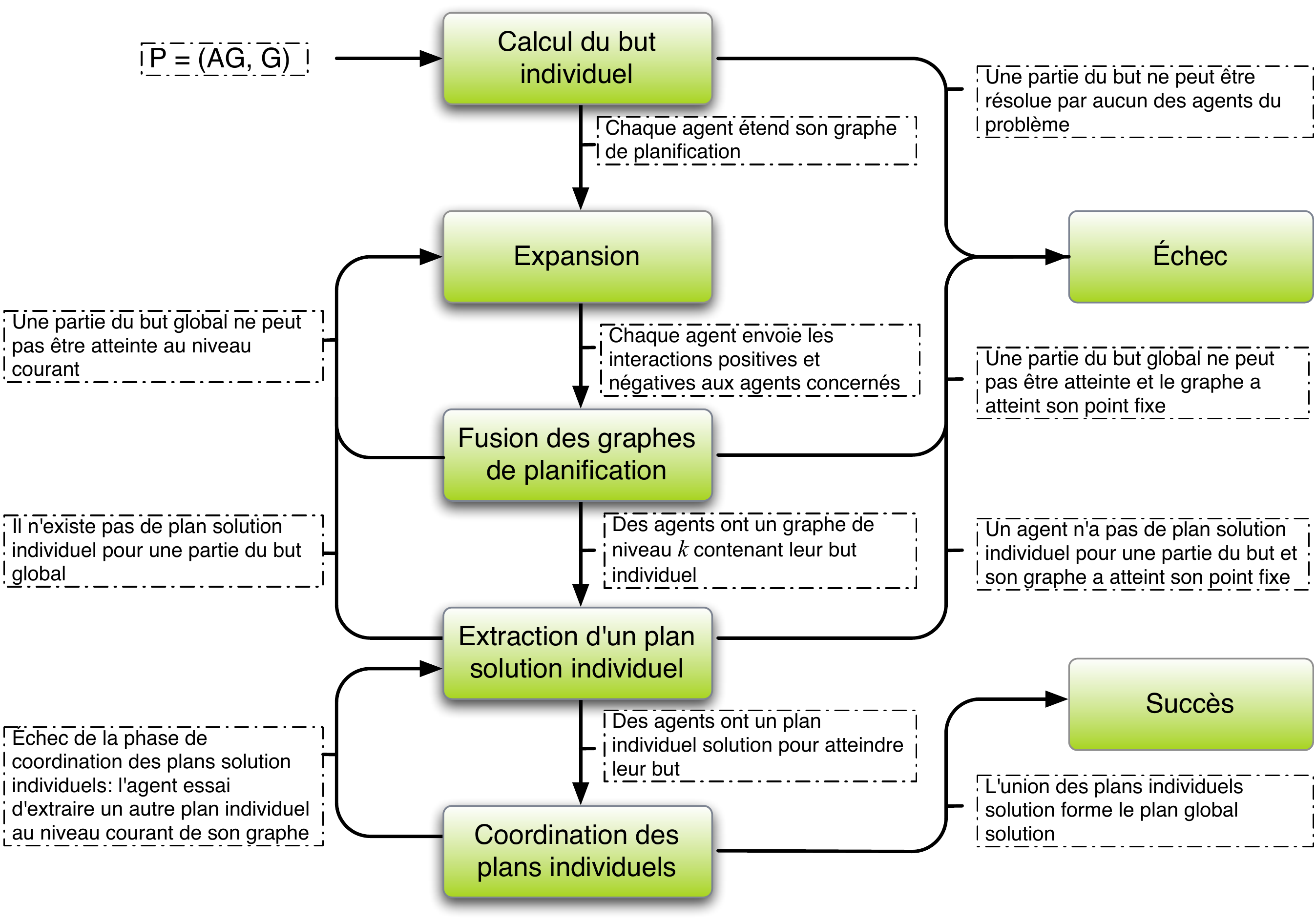}
\end{center}
\caption{Organisation de la procédure de planification distribuée par fusions incrémentales de graphes.}
\label{Fig:DGMP}
\end{figure*}

\operator[Le docker \cons{ag1} charge un conteneur \var{c} dans le camion \var{t} à l'emplacement \cons{l1}]
{\op{load}{\cons{l1},\var{c},\var{t}}}
{\pre{at}{\var{t},\cons{l1}}, \pre{at}{\var{c},\cons{l1}}}
{$\neg$\pre{at}{\var{c},\cons{l1}}, \pre{in}{\var{t},\var{c}}}

\operator[Le docker \cons{ag1} décharge un conteneur \var{c} du camion \var{t} à l'emplacement \cons{l1}]
{\op{unload}{\cons{l1},\var{c},\var{t}}}
{\pre{at}{\var{t},\cons{l1}}, \pre{in}{\var{t},\var{c}}}
{$\neg$\pre{in}{\var{t},\var{c}}, \pre{at}{\var{c},\cons{l1}}}

\operator[Le docker \cons{ag2} charge un conteneur \var{c} dans le camion \var{t} à l'emplacement \cons{l2}]
{\op{load}{\cons{l2},\var{c},\var{t}}}
{\pre{at}{\var{t},\cons{l2}}, \pre{at}{\var{c},\cons{l2}}}
{$\neg$\pre{at}{\var{c},\cons{l2}}, \pre{in}{\var{t},\var{c}}}

\operator[Le docker \cons{ag2} décharge un conteneur \var{c} du camion \var{t} à l'emplacement \cons{l2}]
{\op{unload}{\cons{l2},\var{c},\var{t}}}
{\pre{at}{\var{t},\cons{l2}}, \pre{in}{\var{t},\var{c}}}
{$\neg$\pre{in}{\var{t},\var{c}}, \pre{at}{\var{c},\cons{l2}}}
\operator[Le convoyeur déplace le camion \var{t} de l'emplacement \var{from} à l'emplacement \var{to}]
{\op{move}{\var{t},\var{from},\var{to}}}
{\pre{at}{\var{t},\var{from}}}
{\pre{at}{\var{t},\var{to}}, $\neg$\pre{at}{\var{t},\var{from}}}
\label{Ex:Dockers}
\end{example}

\section{Dynamique du modèle}

L'algorithme de planification distribuée par fusions incrémentales de graphes de planification (cf. {\it fig.~\ref{Fig:DGMP}}) s'articule autour de cinq phases. Il réalise une recherche distribuée en  profondeur d'abord par niveau, découvrant une nouvelle partie de l'espace de recherche à chaque itération. Dans un premier temps, chaque agent calcule les buts individuels associés aux agents du problème puis entre dans une boucle dans laquelle il étend itérativement son graphe de planification, fusionne les interactions negatives et positives provenant de l'activité des autres agents et finalement, effectue une recherche locale d'un plan individuel solution. Si les agents réussissent à extraire un plan solution individuel à partir de leur graphe de planification, ils tentent alors de les coordonner. Dans le cas contraire, chaque agent étend une nouvelle fois son graphe de planification. La boucle d'expansion, de fusion, et de recherche de plans solution individuel se poursuit tant qu'aucun plan solution global n'est trouvé ou que la condition de terminaison sur échec n'est pas vérifiée.

\vspace{-0.5cm}
\subsection{Décomposition du but global}

Tout d'abord, chaque agent calcule les buts individuels associés aux agents du problème ({\it def.}~\ref{Def:But-Individuel}), \ie le sous-ensemble des propositions du but qui co-désignent un effet de ses opérateurs. Si une partie du but global n'apparaît dans aucun des buts individuels des agents, la phase de décomposition échoue, et la synthèse de plans se termine. Cet échec signifie qu'une partie du but global ne peut être atteinte, puisqu'aucun agent ne possède d'opérateurs capables de produire un sous-ensemble des propositions composant le but global. Dans le cas contraire, les agents passent dans la phase d'expansion. Considérons l'exemple \ref{Ex:Dockers}, les buts individuels des agents sont définis de la manière suivante: $g_{\textsf{\scriptsize{ag1}}} = \{\pre{at}{\cons{c2},\cons{l1}}\}$; $g_{\textsf{\scriptsize{ag2}}} = \{\pre{at}{\cons{c1},\cons{l2}}\}$; $g_{\textsf{\scriptsize{ag3}}} = \{\pre{at}{\cons{t1},\cons{l2}},\pre{at}{\cons{t2},\cons{l1}}\}$. Il est important de noter qu'un agent peut ne pas avoir de but individuel, mais être nécessaire à la synthèse de plan. Dans ce cas, l'agent étend malgré tout son graphe pour lui permettre d'évaluer les actions d'aide qu'il peut proposer aux autres agents.

\subsection{Expansion du graphe de planification}

\begin{figure}[!h]

\begin{center}
\subfigure[Graphe de planification de l'agent 1: but individuel non atteint.]{
\includegraphics[scale=0.53]{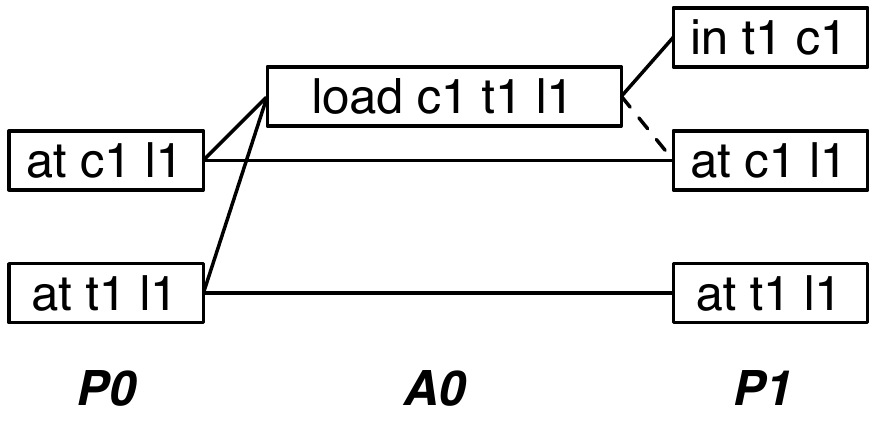}
}
\subfigure[Graphe de planification de l'agent 2: but individuel non atteint.]{
\includegraphics[scale=0.53]{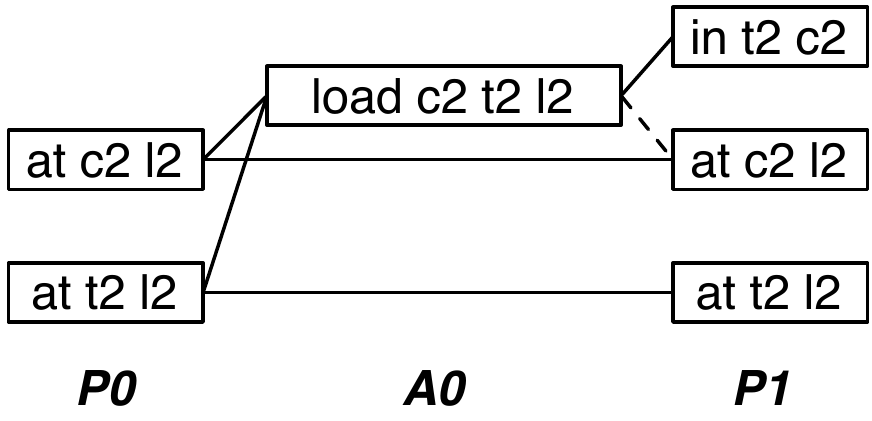}
}
\subfigure[Graphe de planification de l'agent 3: but individuel atteint et exempt d'exclusion mutuelle.]{
\includegraphics[scale=0.53]{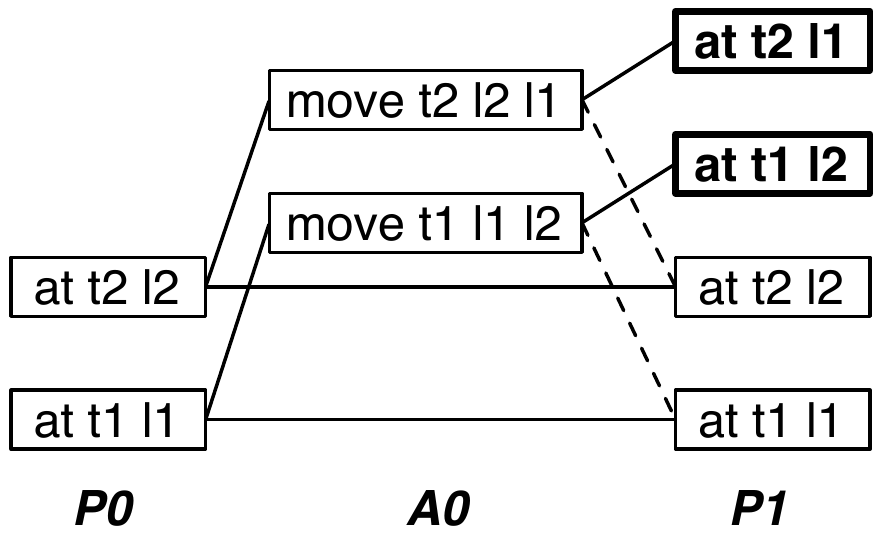}
}
\end{center}
\caption{Graphe de planification des agents de l'exemple \ref{Ex:Dockers}.}
\label{Fig:APG-1}
\end{figure}

Cette phase débute par l'initialisation, par chaque agent, de son graphe de planification à partir de ses croyances. Puis, chaque agent étend son graphe en respectant la définition \ref{Def:Planning-Graph}. Lorsque l'expansion de son graphe est terminée, l'agent envoie aux agents concernés les actions associées à son graphe qui peuvent interférer soit de manière positive ({\it def.}~\ref{Def:Relation-Positive}) soit de manière négative ({\it def.}~\ref{Def:Relation-Negative}). La figure \ref{Fig:APG-1} donne les graphes de planification associés aux agents $\cons{ag}_\cons{1}$ (a), $\cons{ag}_\cons{2}$ (b) and $\cons{ag}_\cons{3}$ (c) après la premier expansion de leur graphe : les boites au niveau $P_i$ représentent des propositions et les boites au niveau $A_i$ des actions; pour simplifier, les relations d'exclusion mutuelles ne sont pas représentées; les lignes pleines modélisent les préconditions et les effets positifs des actions tandis que les lignes pointillées modélisent les effets négatifs; finalement, les boites en gras définissent les propositions but lorsqu'elles sont atteintes sans mutex. Le graphe de $\cons{ag}_\cons{3}$ contient son but individuel exempt d'exclusion mutuelle au niveau 1. En revanche, les graphes de planification de $\cons{ag}_\cons{1}$ et $\cons{ag}_\cons{2}$ ne contiennent par leur but individuel respectif. Considérons maintenant les interactions négatives et positives entre les actions des différents graphes. Par exemple, l'action \pre{move}{\cons{t1},\cons{l1},\cons{l2}} de $\cons{ag}_\cons{3}$ au  niveau 0 menace l'activité de  $\cons{ag}_\cons{1}$ car elle supprime la précondition \pre{at}{\cons{t1},\cons{l1}} qui peut être nécessaire à l'application de l'opérateur \pre{load}{} et \pre{unload}{} de $\cons{ag}_\cons{1}$. Par opposition, cette même action de  $\cons{ag}_\cons{3}$ assiste l'activité de $\cons{ag}_\cons{2}$ puisqu'elle produit l'effet \pre{at}{\cons{t1},\cons{l2}} qui peut être nécessaire au déclenchement des opérateurs \pre{load}{} et \pre{unload}{} de $\cons{ag}_\cons{2}$. La liste complète des interactions entre les actions des agents est donnée par le tableau \ref{Tab:Interactions}.

\begin{table}[!h]
\begin{center}
\begin{tabular}{l||c|c|c}
\textsl{Actions} & \textsl{Agent} & \textsl{Négative} & \textsl{Positive} \\
\hline
\pre{load}{\cons{c1},\cons{t1},\cons{l1}}   & $\cons{ag}_\cons{1}$ & \---                 & $\cons{ag}_\cons{2}$ \\
\pre{load}{\cons{c2},\cons{t2},\cons{l2}}   & $\cons{ag}_\cons{2}$ & \---                 & $\cons{ag}_\cons{1}$ \\
\pre{move}{\cons{t1},\cons{l1},\cons{l2}}   & $\cons{ag}_\cons{3}$ & $\cons{ag}_\cons{1}$ & $\cons{ag}_\cons{2}$ \\
\pre{move}{\cons{t2},\cons{l2},\cons{l1}}   & $\cons{ag}_\cons{3}$ & $\cons{ag}_\cons{2}$ & $\cons{ag}_\cons{1}$ \\
\end{tabular}
\caption{Table des interactions entre agents au niveau 0.}
\label{Tab:Interactions}
\end{center}
\end{table}

\subsection{Fusions des graphes de planification}

Au cours de cette phase, l'agent fusionne les actions pouvant interférer avec son activité provenant des autres agents à son graphe de planification. Autrement dit, l'agent ajoute les actions à son graphe et achève son expansion \ie déclenche toutes actions possibles à partir des nouvelles propositions introduites au cours de l'expansion de son graphe. Le graphe contient désormais les actions provenant des autres agents pouvant interférer de manière positive ou négative avec sa propre activité. Deux cas sont à envisager:
\begin{enumerate}
\item Un agent $\alpha$ nécessaire à la réalisation du but global possède un graphe qui a atteint son point fixe au niveau $i$ et $g_\alpha \not \subseteq P_i$ ou $g_\alpha \in P_i$ et $g_\alpha \cap \mu P_i \not= \emptyset$: la synthèse de plans échoue et l'algorithme se termine. En effet, un sous-ensemble des propositions du but global ne peut plus être atteint.
\item Un agent $\alpha$ nécessaire à la réalisation du but global possède un graphe qui n'a pas atteint son point fixe au niveau $i$ mais $g_\alpha \not \subseteq P_i$ ou $g_\alpha \in P_i$ et $g_\alpha \cap \mu P_i \not= \emptyset$: tous les agents effectuent une nouvelle expansion de leur graphe de planification.
\end{enumerate}

Finalement, si aucune des deux conditions n'est vérifiée, la phase de fusion se termine en garantissant qu'il existe un sous-ensemble des agents du problème capables d'atteindre le but global, et que chaque agent de ce sous-ensemble possède un graphe de planification de niveau $i$, tel que $g_\alpha \subseteq P_i$ et $g_\alpha \cup \mu P_i = \emptyset$.

\begin{figure*}[!]
\begin{center}
\includegraphics[scale=0.55]{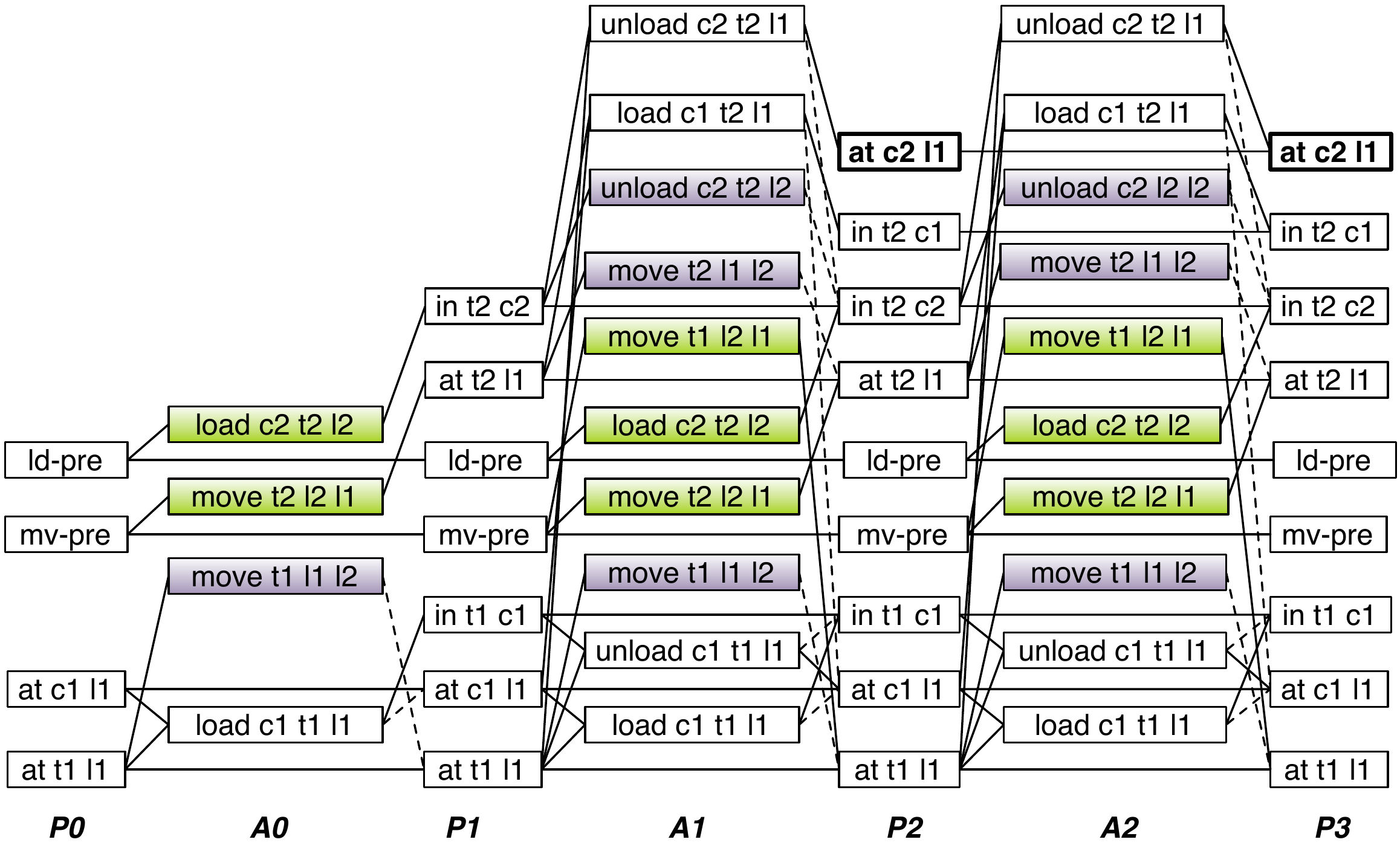}
\includegraphics[scale=0.55]{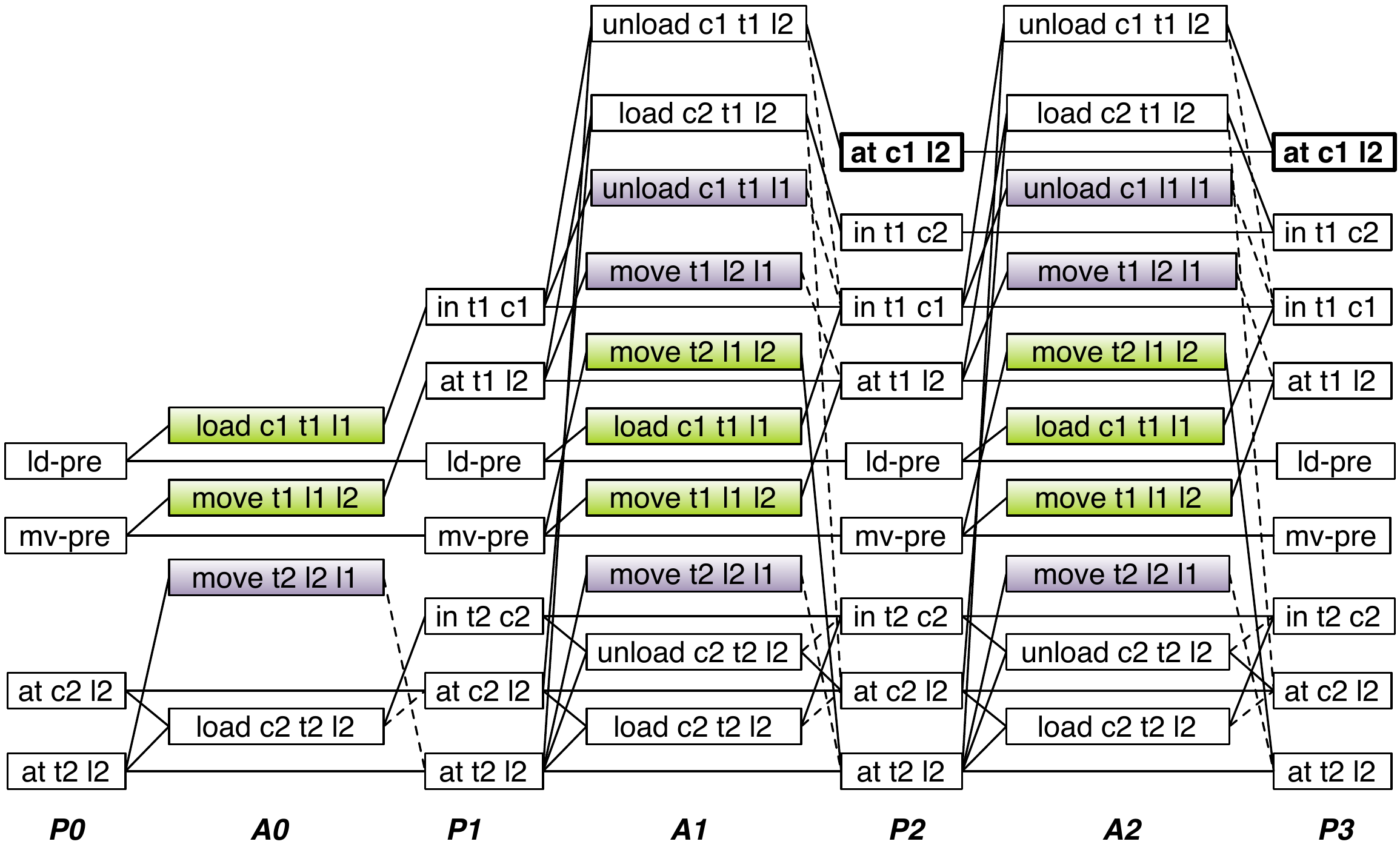}
\includegraphics[scale=0.55]{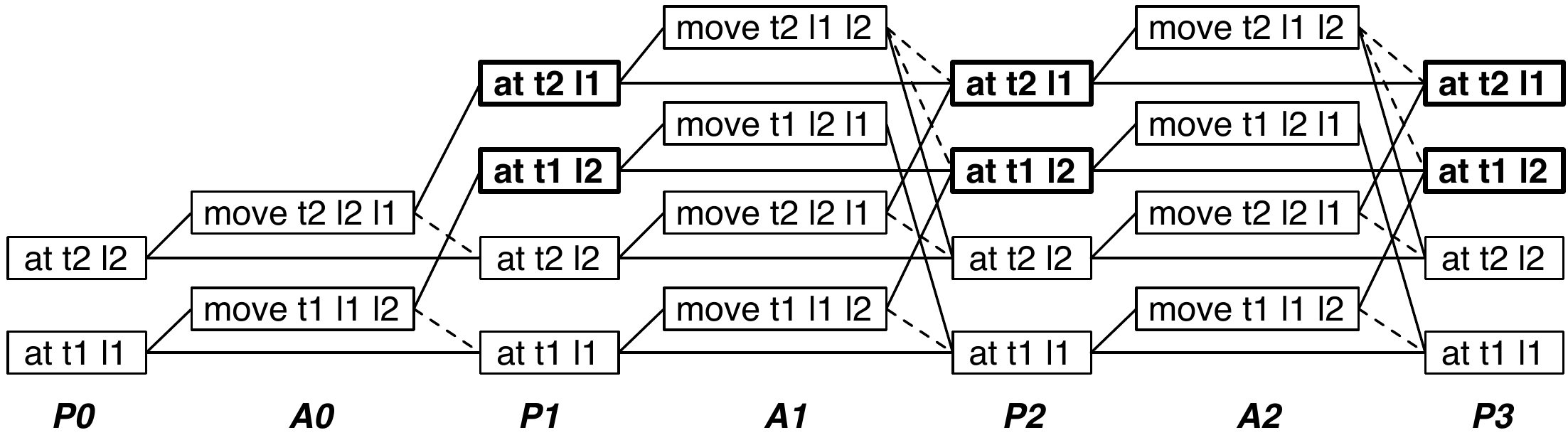}
\end{center}
\caption{Graphes de planification des agents de l'exemple des dockers après la phase d'expansion et de fusion.}
\label{Fig:Dockers-Graphs}
\end{figure*}

À titre d'illustration, nous donnons à la figure \ref{Fig:Dockers-Graphs} les graphes de planification des trois agents de l'exemple \ref{Ex:Dockers} après trois itérations d'expansion et de fusion. Afin de limiter l'ajout de propositions inutiles au niveau du graphe de planification pour ne pas pénaliser la recherche de plans, seules les propositions apparaissant dans la description des opérateurs d'un agent sont ajoutées. Par exemple, l'ajout de l'action \op{move}{\cons{t2},\cons{l2},\cons{l1}} de l'agent \cons{ag3} dans le graphe de planification de l'agent \cons{ag1} ne produira qu'un seul effet, \pre{at}{\cons{t2},\cons{l1}}, au niveau 1 du graphe de \cons{ag1}. Le même principe s'applique également aux préconditions des actions ajoutées. Toutefois, si l'action n'est pas liée par au moins une précondition au niveau propositionnel (c'est le cas de l'action \op{move}{\cons{t2},\cons{l2},\cons{l1}}), nous ajoutons une précondition fictive pour garantir l'existence d'un chemin de $P_0$ à $P_i$, et ainsi permettre l'extraction d'un plan individuel solution.

\subsection{Extraction d'un plan individuel}

L'extraction d'un plan solution est effectuée en s'appuyant sur une technique de satisfaction de contraintes proposée par \cite{kambhampati:00}. Cette technique présente deux principaux avantages: d'une part, elle améliore de manière significative les temps d'extraction des plans solution et d'autre part, elle peut être facilement modifiée pour extraire un plan solution individuel intégrant les contraintes provenant des autres agents. L'extraction débute par l'encodage du graphe de planification sous la forme d'un CSP. Chaque proposition $p$ à un niveau $i$ du graphe de planification est assimilée à une variable CSP. L'ensemble des actions qui produisent $p$ au niveau $n$ constitue son domaine auquel on ajoute une valeur $\perp$ nécessaire à l'activation des actions dans le graphe. Les relations d'exclusion correspondent aux contraintes du CSP. Lorsque deux actions $a_1$ et $a_2$ sont mutuellement exclusives, alors pour toutes les paires de propositions $p_1$ et $p_2$ telles que $a_1$ peut produire $p_1$ et $a_2$ respectivement $p_2$, la contrainte $(p_1=a_1) \Rightarrow (p_2\not=a_2)$ est ajoutée au CSP. Lorsque deux propositions $p_1$ et $p_2$ sont mutuellement exclusives, on le traduit par une contrainte $\neg(p_1 \not= \perp) \wedge (p_2 \not= \perp)$. L'assignation des valeurs aux variables est dynamique car chaque assignation, à un niveau donné, active des variables au niveau précèdent en respectant la procédure classique d'extraction de graphplan. Initialement, seules les variables représentant le but individuel de l'agent sont actives. L'activation dynamique se traduit par des contraintes d'activation au niveau du CSP. Autrement dit, lorsque une variable, représentant une proposition, prend une certaine valeur, \ie est produite par une action, alors d'autres variables, qui correspondent aux préconditions de l'action choisie, deviennent actives. L'extraction d'un plan solution individuel à partir du graphe de planification de l'agent est réalisée en résolvant le CSP. En partant des variables actives initialement, \ie le but individuel de l'agent, la procédure d'extraction cherche à assigner une valeur ou une action, à chaque variable ou proposition, pour satisfaire l'ensemble des contraintes, \ie des exclusions mutuelles. Si la résolution du CSP réussie, l'agent a extrait un plan solution individuel. Dans le cas contraire, l'agent n'est pas capable de produire un plan individuel solution. Deux cas doivent être considérés:
\begin{enumerate}
\item Un agent nécessaire à la réalisation du but global ne parvient pas à extraire un plan solution individuel même en poursuivant l'expansion de son graphe, l'algorithme se termine. Une partie du but global ne peut être démontrée.
\item Un agent nécessaire à la réalisation du but global ne parvient pas à extraire un plan solution individuel mais peut espérer en trouver un en effectuant une nouvelle expansion de son graphe. Tous les agents effectuent une nouvelle expansion de leur graphe de planification.
\end{enumerate}

Finalement, la phase d'extraction se termine en garantissant qu'il existe un sous-ensemble des agents du problème capables d'atteindre le but global, et possèdant un plan individuel solution.

\subsection{Coordination des plans individuels}

Il s'agit maintenant de vérifier la compatibilité des plans individuels dans le contexte multi-agent. Rappelons que lors de la phase de fusion, les agents ont ajouté dans leur graphe les actions provenant d'autres agents et pouvant interférer avec leur propre activité. Par conséquent, un plan individuel solution peut contenir des actions qui doivent être exécutées par d'autres agents. Autrement dit, un plan solution individuel est un plan conditionnel \ie un plan exécutable si certaines contraintes sont vérifiées. Dans notre cas, ces contraintes sont définis par des couples $(a, i)$ ou $a$ représente une action et $i$ le niveau où elle doit être exécutée. Considérons l'exemple des dockers et leur graphe de planification (cf. {\it fig.} \ref{Fig:Dockers-Graphs}). Les agents \cons{ag1} et \cons{ag2} peuvent extraire respectivement un plan solution individuel au niveau 2:
\begin{eqnarray*}
\pi_{\textsf{\scriptsize{ag1}}} & = & \langle \pre{load}{\cons{c2},\cons{t2},\cons{l2}}, \pre{move}{\cons{t2},\cons{l2},\cons{l1}}, \pre{unload}{\cons{c2},\cons{t2},\cons{l1}} \rangle \\
\pi_{\textsf{\scriptsize{ag2}}} & = & \langle \pre{load}{\cons{c1},\cons{t1},\cons{l1}}, \pre{move}{\cons{t1},\cons{l1},\cons{l2}}, \pre{unload}{\cons{c1},\cons{t1},\cons{l2}} \rangle
\end{eqnarray*}
Le plan $\pi_{\textsf{\scriptsize{ag1}}}$ est valide si le plan individuel de \cons{ag2} respecte la contrainte $(\pre{load}{\cons{c2},\cons{t2},\cons{l2}}, 0)$ et celui de \cons{ag3} la contrainte $(\pre{move}{\cons{t2},\cons{l2},\cons{l1}}, 1)$. De manière symétrique, $\pi_{\textsf{\scriptsize{ag2}}}$ est valide si le plan individuel de \cons{ag2} respecte $(\pre{load}{\cons{c1},\cons{t1},\cons{l1}}, 0)$ et celui de \cons{ag3} la contrainte $(\pre{move}{\cons{t1},\cons{l1},\cons{l2}}, 1)$.
De son côté, l'agent convoyeur \cons{ag3} peut potentiellement extraire plusieurs plans individuels solution dont aucun n'implique des actions provenant des autres agents:
\begin{footnotesize}
\begin{eqnarray*}
\pi_{\textsf{\scriptsize{ag3}}} & = & \langle \{\pre{move}{\cons{t2},\cons{l2},\cons{l1}}, \pre{move}{\cons{t1},\cons{l1},\cons{l2}}\}, \pre{no-op}{}, \pre{no-op}{}  \rangle \\
\pi_{\textsf{\scriptsize{ag3}}}' & = & \langle \pre{move}{\cons{t2},\cons{l2},\cons{l1}}, \pre{move}{\cons{t1},\cons{l1},\cons{l2}}, \pre{no-op}{}  \rangle \\
\pi_{\textsf{\scriptsize{ag3}}}'' & = & \ldots
\end{eqnarray*}
\end{footnotesize}
Les agents débutent la phase de coordination par l'échange des contraintes de leur plan individuel, et tentent ensuite d'intégrer les contraintes reçues à leur plan. L'intégration des contraintes repose sur le principe de moindre engagement. Tout d'abord, l'agent teste si les contraintes peuvent être ajoutées directement à son plan individuel solution. Ce teste ne nécessite aucune replanification, puisqu'il suffit de vérifier que l'action à ajouter n'est pas mutuellement exclusive avec une action déjà présente dans le graphe de planification de l'agent au même niveau. Dans notre exemple, ce mécanisme de coordination sera utilisé par les agents \cons{ag1} et \cons{ag2} pour prendre respectivement en compte les contraintes $(\pre{load}{\cons{c2},\cons{t2},\cons{l2}}, 0)$ et $(\pre{load}{\cons{c1},\cons{t1},\cons{l1}}, 0)$. Si ce premier mécanisme échoue, l'agent tente alors d'extraire un nouveau plan individuel solution en incluant les contraintes. En d'autres termes, il propage les contraintes dans le graphe de contraintes représentant son graphe de planification, et effectue l'extraction d'un nouveau plan individuel solution. Ce mécanisme sera utilisé par l'agent \cons{ag3} pour intégrer les contraintes $(\pre{move}{\cons{t1},\cons{l1},\cons{l2}}, 1)$ et $(\pre{move}{\cons{t2},\cons{l2},\cons{l1}}, 1)$ provenant des agents \cons{ag1} et \cons{ag2}. Au final, les plans individuels solutions de l'exemple \ref{Ex:Dockers} obtenus après la phase de coordination sont les suivants:
\begin{eqnarray*}
\pi_{\textsf{\scriptsize{ag1}}} & = & \langle \pre{load}{\cons{c1},\cons{t1},\cons{l1}}, \pre{no-op}{}, \pre{unload}{\cons{c2},\cons{t2},\cons{l1}} \rangle \\
\pi_{\textsf{\scriptsize{ag2}}} & = & \langle \pre{load}{\cons{c2},\cons{t2},\cons{l2}}, \pre{no-op}{}, \pre{unload}{\cons{c1},\cons{t1},\cons{l2}} \rangle\\
\pi_{\textsf{\scriptsize{ag3}}} & = & \langle \pre{no-op}{}, \{\pre{move}{\cons{t1},\cons{l1},\cons{l2}}, \pre{move}{\cons{t2},\cons{l2},\cons{l1}}\}, \pre{no-op}{} \rangle
\end{eqnarray*}
Finalement, si ce second mécanisme de coordination échoue, les agents retournent dans la phase d'extraction de plans individuel. L'algorithme de fusions incrémentales de plans peut alors se terminer (si aucun autre plan individuel ne peut être extrait) ou encore nécessité une nouvelle expansion des graphes de planification des agents.

\section{Conclusion}

Dans cet article, nous avons proposé un modèle générique et original pour la synthèse distribuée de plans par un groupe d'agents, appelé {\em planification distribuée par fusions incrémentales de graphes}. Le modèle semble correct et complet, et unifie de manière élégante les différentes phases de la planification distribuée au sein d'un même processus. En outre, il permet aux agents de limiter les interactions négatives entre leur plan individuel, mais également de prendre en compte leurs interactions positives, \ie d'aide ou d'assistance, au plus tôt, \ie avant l'extraction des plans individuels. Notre approche est actuellement en cours d'implantation en s'appuyant sur les librairies PDDL4J\footnote{\url{https://github.com/pellierd/pddl4j}} et Choco\footnote{\url{http://choco-solver.net/}}. Chose intéressante, les premiers résultats expérimentaux semblent indiquer que la complexité n'est pas dépendante du nombre d'agents présents dans le problème, mais uniquement du couplage entre les activités des agents, \ie des interactions positives. Plusieurs pistes pour la poursuite de ce travail sont envisagées:
\begin{itemize}
\item la prise en compte de la robustesse des plans. Par robustesse, nous entendons la capacité d'un plan à tolérer plus ou moins des aléas d'exécution ou des croyances erronées sur l'état du monde. Supposons que nous soyons capables de construire un faisceau de plans solutions, \ie un plan contenant différentes alternatives. La question qui se pose alors est la suivante : comment choisir les plans individuels les plus robustes minimisant ainsi le risque d'échec à l'exécution ?
\item l'extraction de plans temporels flottants, \ie de plans dans lesquels les actions sont représentées par des intervalles pouvant être décalés dans le temps. Bien que la causalité soit suffisante pour un grand nombre d'applications, la prise en compte des aspects temporels est parfois nécessaire. La notion de plans flottants apparaît particulièrement intéressante dans un contexte multi-agent, car ils possèdent une grande flexibilité qui peut être mise à profit au cours de la phase de coordination et d'exécution.
\end{itemize}

\newcommand{\etalchar}[1]{$^{#1}$}


\begin{thebibliography}{BNBYF{\etalchar{+}}01}

\bibitem[AFH{\etalchar{+}}98]{alami:98}
R.~Alami, S.~Fleury, M.~Herrb, F.~Ingrand, and F.~Robert.
\newblock Multi robot cooperation in the {\sc martha} project.
\newblock {\em IEEE Robotics and Automation Magazine}, 5(1):36--47, 1998.

\bibitem[BF97]{blum:97}
A.~Blum and M.~Furst.
\newblock {F}ast {P}lanning {T}hrough {P}lanning {G}raph {A}nalysis.
\newblock {\em Artificial Intelligence}, 90(1-2):281--300, 1997.

\bibitem[BNBYF{\etalchar{+}}01]{bar-noy:01}
A.~Bar-Noy, R.~Bar-Yehuda, A.~Freund, J.~Naor, and B.~Schieber.
\newblock A unified approach to approximating resource allocation and
  scheduling.
\newblock {\em Journal of Association for Computing Machinery},
  48(5):1069--1090, 2001.

\bibitem[CB03]{clement:03}
B.~Clement and A.~Barrett.
\newblock Continual coordination through shared activities.
\newblock In {\em Proceedings of the International Conference on Autonomous
  Agent and Muti-Agent Systems}, pages 57--67, 2003.

\bibitem[CD05]{cox:05}
J.~Cox and E.~Durfee.
\newblock An efficient algorithm for multiagent plan coordination.
\newblock In {\em Proceedings of the International Joint Conference on
  Autonomous Agents and Multi-Agent Systems}, pages 828--835, New York, NY,
  USA, 2005. ACM Press.

\bibitem[FGLS06]{fox:06}
M.~Fox, A.~Gerevini, D.~Long, and I.~Serina.
\newblock Plan stability: Replanning versus plan repair.
\newblock In {\em Proceedings of the International Conference on Planning and
  Scheduling}, California, USA, 2006. AAAI Press.

\bibitem[IM02]{iwen:02}
M.~Iwen and A.~D. Mali.
\newblock Automatic problem decomposition for distributed planning.
\newblock In {\em Proceedings of the International Conference on Artificial
  Intelligence}, pages 411--417, 2002.

\bibitem[Kam00]{kambhampati:00}
S.~Kambhampati.
\newblock Planning graph as a (dynamic) {CSP}: Exploiting {EBL}, {DDB} and
  other {CSP} search techniques in graphplan.
\newblock {\em Journal of Artificial Intelligence Research}, 12(1):1--34, 2000.

\bibitem[LDW{\etalchar{+}}04]{lesser:04}
V.~Lesser, K.~Decker, T.~Wagner, N.~Carver, A.~Garvey, B.~Horling, D.~Neiman,
  R.~Podorozhny, M.~NagendraPrasad, A.~Raja, R.~Vincent, P.~Xuan, and X.Q.
  Zhang.
\newblock Evolution of the gpgp/taems domain-independent coordination
  framework.
\newblock In {\em Proceedings of the International Conference on Autonomous
  Agents and Multi-agent Systems}, pages 87--143. Kluwer Academic Publishers,
  2004.

\bibitem[NAI{\etalchar{+}}03]{nau:03}
D.~Nau, T.~Au, O.~Ilghami, U.~Kuter, W.~Murdock, D.~Wu, and Y.~Yaman.
\newblock {SHOP2}: {A}n {HTN} {P}lanning {S}ystem.
\newblock {\em Journal of Artificial Intelligence Research}, 20(1):379--404,
  2003.

\bibitem[PW92]{penberthy:92}
J.~Penberthy and D.~Weld.
\newblock {UCPO}: A sound, complete, partial order planner for {\sc adl}.
\newblock In C.~Rich B.~Nebel and W.~Swartout, editors, {\em Proceedings of the
  International Conference on Principles of Knowledge Representation and
  Reasoning}, pages 103--114. Morgan Kaufmann Publishers, 1992.

\bibitem[ST95]{shoham:95}
Y.~Shoham and M.~Tennenholtz.
\newblock On social laws for artificial agents societies: off-line design.
\newblock {\em Artificial Intelligence}, 73(1--2):231--252, 1995.

\bibitem[TBdWW02]{tonino:02}
H.~Tonino, A.~Bos, M.~de~Weerdt, and C.~Witteveen.
\newblock Plan coordination by revision in collective agent-based systems.
\newblock {\em Artificial Intelligence}, 142(2):121--145, 2002.

\bibitem[TJ99]{tambe:99}
M.~Tambe and H.~Jung.
\newblock The benefits of arguing in a team.
\newblock {\em Artificial Intelligence Magazine}, 20(4):85--92, 1999.

\bibitem[WPSEN03]{wu:03}
D.~Wu, B.~Parsia, J.~Sirin~E, and D.~Nau.
\newblock Automating daml-s web services composition using shop2.
\newblock In {\em Proceedings of International Semantic Web Conference}, 2003.

\bibitem[YS03]{younes:03}
H.~Younes and R.~Simmons.
\newblock {VHPOP}: Versatile heuristic partial order planner.
\newblock {\em Journal of Artificial Intelligence Research}, 20(1):405--430,
  2003.

\bibitem[ZR90]{zlotkin:90}
G.~Zlotkin and J.~Rosenschein.
\newblock Negotiation and conflict resolution in non-cooperative domains.
\newblock In {\em Proceedings of the American National Conference on Artificial
  Intelligence}, pages 100--105, Boston, Massachusetts, 1990.

\end{thebibliography}
\end{document}